\documentclass[11pt,a4paper]{article}
\usepackage[hyperref]{acl2019}
\usepackage{times}
\usepackage{latexsym}

\usepackage{url}

\aclfinalcopy

\usepackage[colorinlistoftodos]{todonotes}
\usepackage{booktabs}
\usepackage{amsmath}
\usepackage{flexisym}
\usepackage{dcolumn}
\usepackage{kotex}
\usepackage{color}
\usepackage{hhline}
\usepackage{verbatim}
\usepackage{multirow}
\usepackage{amssymb}
\usepackage{rotating}
\usepackage{subfigure}
\usepackage{dblfloatfix}
\usepackage{enumitem}
\newcolumntype{L}[1]{>{\raggedright\let\newline\\\arraybackslash\hspace{0pt}}m{#1}}
\newcolumntype{C}[1]{>{\centering\let\newline\\\arraybackslash\hspace{0pt}}m{#1}}
\newcolumntype{R}[1]{>{\raggedleft\let\newline\\\arraybackslash\hspace{0pt}}m{#1}}

\def\BigRoman{\uppercase\expandafter{\romannumeral\number\count 255 }}
\def\Romannumeral{\afterassignment\BigRoman\count255=}

\title{Surf at MEDIQA 2019: Improving Performance of \\ Natural Language Inference in the Clinical Domain \\ by Adopting Pre-trained Language Model}

\author{Jiin Nam \\
  AI Core Team \\
  Samsung Research \\
  Seoul, Korea \\
  \texttt{jiin.nam@samsung.com} \\\And
  Seunghyun Yoon\\
  Dept. ECE \\
  Seoul National University \\
  Seoul, Korea \\
  \texttt{mysmilesh@snu.ac.kr} \\\And
  Kyomin Jung\\
  Dept. ECE \\
  Seoul National University \\
  Seoul, Korea \\
  \texttt{kjung@snu.ac.kr} \\}

\date{}

\begin{document}
\maketitle
\begin{abstract}
While deep learning techniques have shown promising results in many natural language processing (NLP) tasks, it has not been widely applied to the clinical domain.
The lack of large datasets and the pervasive use of domain-specific language (i.e. abbreviations and acronyms) in the clinical domain causes slower progress in NLP tasks than that of the general NLP tasks.
To fill this gap, we employ word/subword-level based models that adopt large-scale data-driven methods such as pre-trained language models and transfer learning in analyzing text for the clinical domain.
Empirical results demonstrate the superiority of the proposed methods by achieving 90.6\% accuracy in medical domain natural language inference task.
Furthermore, we inspect the independent strengths of the proposed approaches in quantitative and qualitative manners. This analysis will help researchers to select necessary components in building models for the medical domain.


\end{abstract}

\section{Introduction}
\label{sec:introduction}
Natural language processing (NLP) has broadened its applications rapidly in recent years such as question answering, neural machine translation, natural language inference, and other language-related tasks.
Unlike other tasks in NLP area, the lack of large labeled datasets and restricted access in the clinical domain have discouraged active participation of NLP researchers for this domain~\cite{romanov2018lessons}.
Furthermore, the pervasive use of abbreviations and acronyms in the clinical domain causes the difficulty of text normalization and makes the related tasks more difficult~\cite{pakhomov2002semi}.

In building NLP models, a word embedding layer that transforms a sequence of tokens in text into a vector representation is considered as one of the fundamental components.
In recent studies, it has been shown that the pre-trained language models by using a huge diversity of corpus (i.e. BERT~\cite{devlin2018bert} and ELMo~\cite{peters2018deep}) generate deep contextualized word representations.
These methods have shown to be very effective for improving the performance of a wide range of NLP tasks by enabling better text understanding and have become a crucial part of the tasks since they have published.



\begin{table*}[t]
\centering
\small
\begin{tabular}{C{0.05\columnwidth}L{0.85\columnwidth}L{0.65\columnwidth}C{0.24\columnwidth}}
\toprule
\textbf{\#}& \textbf{Premise} & \textbf{Hypothesis} & \textbf{Label}\\

\midrule
1&She was treated with Magnesium Sulfate, Labetalol, Hydralazine and bedrest as well as betamethasone.& The patient is pregnant.& entailment \\

\midrule
2&Denied headache, sinus tenderness, rhinorrhea or congestion.& Patient has history of dysphagia& contradiction \\

\midrule
3&Type II Diabetes Mellitus 3.&The patient does not require insulin.&neutral \\

\midrule
4&Ruled in for NSTEMI with troponin 0.11.&The patient has myocardial ischemia.&entailment \\

\midrule
5&Her CXR was clear and it did not appear she had an infection.&Chest x-ray showed infiltrates&contradiction\\

\midrule
6&CHF, EF 55\% 6.&complains of shortness of breath&neutral \\

\bottomrule
\end{tabular}
\caption{Examples from the development set of MedNLI.
}
\label{table:mednli_examples}
\end{table*}

To stimulate the research in the clinical domain, researchers have further investigated to transform the pre-trained language models from general purpose version into the medical domain-specific version.
\citet{lee2019biobert} propose BioBERT that utilizes large-scale bio-medical corpora, PubMed abstracts (PubMed) and PubMed Central full-text articles (PMC), to obtain a medical domain specific language representation through fine-tuning the BERT. Similarity, a PubMed-ELMo\footnote{https://allennlp.org/elmo}, trained with medical domain corpus, is released as one of the contributed ELMo models for medical domain researchers. However, these models are not yet fully explored in medical domain tasks.

Besides these general efforts in building better word representations,~\citet{romanov2018lessons} introduce a large and publicly available natural language inference (NLI) dataset, called MedNLI, for the medical domain (see table~\ref{table:mednli_examples}).
Considering the expensive annotation cost of medical text due to the sparsity of the clinical-domain experts, the medical NLI task plays an import role in boosting existing datasets for medical question answering systems by retrieving similar questions that are already answered by human experts.
Along with this effort, ACL-BioNLP 2019 committee announced a shared task, NLI for the medical domain, motivated by a need to develop relevant methods, techniques and gold standards for inference and entailment~\cite{MEDIQA2019}. The newly released dataset is larger in size than that of any other previous medical domain NLI dataset, however, it is still not enough to train complicated neural network based models. 

To fill this gap, we propose a combination approach of NLP models and machine learning methods to tackle the medical domain NLI task. 
Our contributions are summarized as follows:



\begin{itemize}
	\item We adopt the pre-trained language models (BioBERT, PubMed-ELMo) to overcome the shortage of training data which is a common problem in the clinical domain.
	\item We apply the transfer learning method with two general domain NLI datasets and show that a source task in a domain can benefit learning a target task in a different domain.
	\item We show the independent strengths of the proposed approaches in quantitative and qualitative manners. This analysis will help researchers to select necessary components in building models for the clinical domain.
	
\end{itemize}


\section{Related Work}
\label{sec:realated_work}
Researchers have investigated NLI tasks. Most of the works employed a recurrent neural network to encode each pair of sentences and to compute the similarity between them~\cite{conneau2017supervised,subramanian2018learning}.
Recently,~\citet{liu2019multi} proposed multi-task learning for natural language tasks and achieved the best results on NLI tasks.
In the medical domain, \citet{romanov2018lessons} adopted the ESIM~\cite{chen2017enhanced} model to the MedNLI task. 
The ESIM model employs two bidirectional LSTM to encode each sentence independently and to calculate a matching score between the sentences by using alignment and pooling methods. They also applied transfer learning with SNLI~\cite{bowman2015large} and MNLI~\cite{williams2018broad} datasets to improve model performance in the MedNLI task.

Recently, pre-trained language models were proposed~\cite{peters2018deep,devlin2018bert}. The multi-task benchmark for natural language understanding~\cite{wang2018glue} has shown that these pre-trained language models brought additional performance gain by providing deep contextualized word representations. Upon this success, researchers further extended previous pre-trained language models to medical domain-specific versions such as BioBERT~\cite{lee2019biobert} and PubMed-ELMo~\cite{pubmed_emlo2018}.

However, none of these researches directly applied the pre-trained language models of the medical domain to the MedNLI task.

\section{Dataset and Problem}
\label{sec:dataset}

MedNLI~\cite{romanov2018lessons}, a large publicly available and expert annotated dataset, has been recently published for the MEDIQA 2019 shared task. 
This dataset comprises of tuples ${<}P,H,Y{>}$ where: \textit{P} and \textit{H} are a clinical sentence pair, (premise and hypothesis, respectively); \textit{Y} indicates whether a given hypothesis can be
inferred from a given premise. In particular, \textit{Y} is categorized as one of three classes: ``\textit{entailment}", ``\textit{contradiction}", and ``\textit{neutral}". Table~\ref{table:mednli_examples} shows examples of the MedNLI dataset. A total of 14,049 pairs, (11,232, 1,395, 1,422 for training, development, and test, respectively), are created based on the past medical history section of MIMIC-{\Romannumeral 3}~\cite{johnson2016mimic}.


In this research, we are interested in building a model that classifies the given sentence pair into the corresponding category. First, we consider a \textbf{point-wise} approach that classifies each pair of data independently into one of the three classes. Next, we re-organize the dataset into the set of a list that contains one of each class sentence pair. Then we apply \textbf{list-wise} classification that classifies three sentence pair into each ``\textit{entailment}", ``\textit{contradiction}", and ``\textit{neutral}" class exclusively.

\begin{figure}[t]
\centering
\includegraphics[width=0.85\columnwidth]{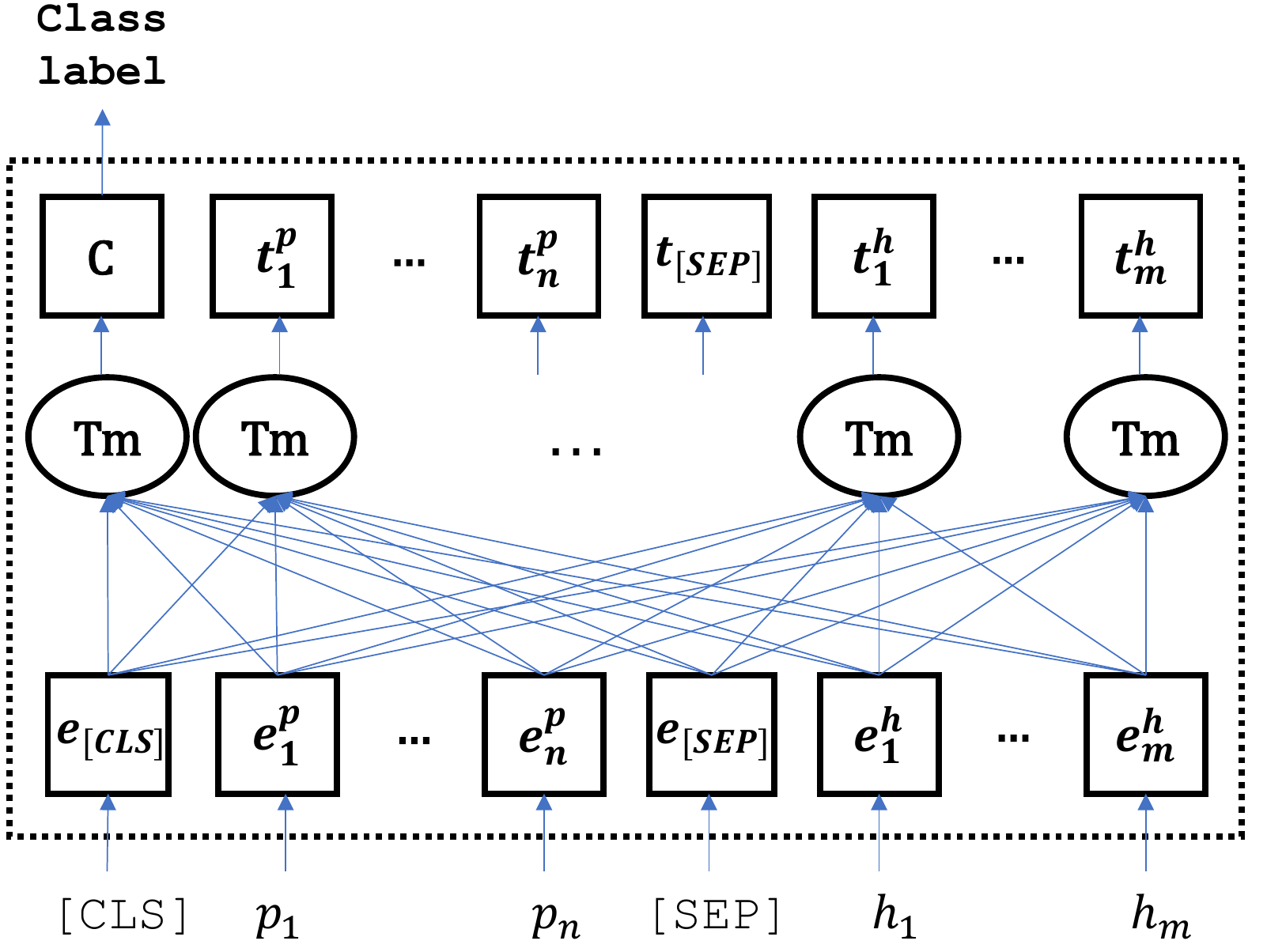}
\caption{
Overview of the BERT model.
}
\label{fig:bert}
\end{figure}

\section{Methods}
\label{sec:methods}
As the size of the MedNLI dataset is limited to train the whole weight parameters in complicated neural network based models, we first choose a BERT~\cite{devlin2018bert} based model that provides pre-trained model parameters from a large corpus. 
To further explore the performance of modern neural network based models, we extend the compare aggregate model~\cite{wang2016compare} with another type of pre-trained word-level embedding, ELMo~\cite{peters2018deep}.
Additionally, we apply transfer learning from similar NLI tasks~\cite{bowman2015large,williams2018broad}, and we try to expand medical abbreviations to deal with the general problem in the medical domain.


\subsection{BioBERT}
As a baseline model, we choose BioBERT~\cite{lee2019biobert} since MedNLI is a bio-domain specific NLI task. 
It shows strength in understanding medical domain text as it is fine-tuned with bio-datasets such as PubMed and PMC.
The BioBERT adopts the same architecture as BERT, as shown in figure~\ref{fig:bert}, that takes WordPiece embeddings from textual input and generates a language representation using a transformer model~\cite{vaswani2017attention}.

\vspace*{1mm}
\noindent\textbf{WordPiece embedding: }
BioBERT utilizes the WordPiece dictionary of BERT generated from general domain corpus. Each premise $\textbf{P}$ and hypothesis $\textbf{H}$ turn into sub-word embeddings, $\textbf{E}^P\,{\in}\,\mathbb{R}^{n\times d_e}$~and~$\textbf{E}^H\,{\in}\,\mathbb{R}^{m\times d_e}$, using the dictionary where $d_e$ is a dimension of sub-word embedding vectors and $n$ and $m$ are the length of the sequences of \textbf{P} and \textbf{H}, respectively.
\begin{equation}
\begin{aligned}
& \textbf{E}^P=\text{WordPiece\_embedding}(\textbf{P}), \\
& \textbf{E}^H=\text{WordPiece\_embedding}(\textbf{H}).
\end{aligned}
\label{eq:wordpiece_embedding}
\end{equation}
BioBERT adds the special classification embedding ``[CLS]" as the first token of every sentence and separates $\textbf{E}^P$ and $\textbf{E}^H$ with a special token ``[SEP]".
The final input representation fed to transformer blocks is the sum of the token embeddings ($\textbf{E}^T$), position embeddings ($\textbf{E}^{Po}$), and segmentation embeddings ($\textbf{E}^S$) as follow.
\begin{equation}
\begin{aligned}
& \textbf{E} = \textbf{E}^T + \textbf{E}^{Po}+ \textbf{E}^S, \\
& \textbf{E}^T=[\textbf{E}_{[CLS]}, \; \textbf{E}^P, \; \textbf{E}_{[SEP]}, \; \textbf{E}^H].
\end{aligned}
\label{eq:final_embeddings}
\end{equation}

\vspace*{1mm}
\noindent\textbf{Transformer encoder: }
The transformer encoder consists of multiple transformer blocks. Each block uses Multi-Head Attention (MHA) generating $h$ different attentions. All the attention heads calculated with different weights are concatenated. A linear layer with a weight matrix $\textbf{W}^H{\in}\mathbb{R}^{(h\times d_v)\times d_e}$ computes the MHA ($\mathbb{R}^{\text{input\_length}\times d_e}$) with the concatenated attention heads as follows:

\begin{equation}
\begin{aligned}
&\text{MHA}\textbf{(Q,\,K,\,V)}=(\text{concat}\{hd_1, .., hd_n\})\textbf{W}^H, \\
& hd_i=\text{attn}(\textbf{Q}_i,\,\textbf{K}_i,\,\textbf{V}_i),\\
& \text{attn}(\textbf{Q}_i,\,\textbf{K}_i,\,\textbf{V}_i)=\text{softmax}(\frac{\textbf{Q}_i\textbf{K}_i^T}{\sqrt{d_k}})\textbf{V}_i,
\end{aligned}
\label{eq:transformer}
\end{equation}
where $Q=[Q_1, ..., Q_h], Q_i \in \mathbb{R}^{n\times \frac{d_e}{h}},$\\
$K=[K_1, ..., K_h], K_i \in \mathbb{R}^{n\times \frac{d_e}{h}},$\\
$V=[V_1, ..., V_h], V_i \in \mathbb{R}^{n\times \frac{d_e}{h}}.$

\begin{figure}[t]
\centering
\includegraphics[width=0.85\columnwidth]{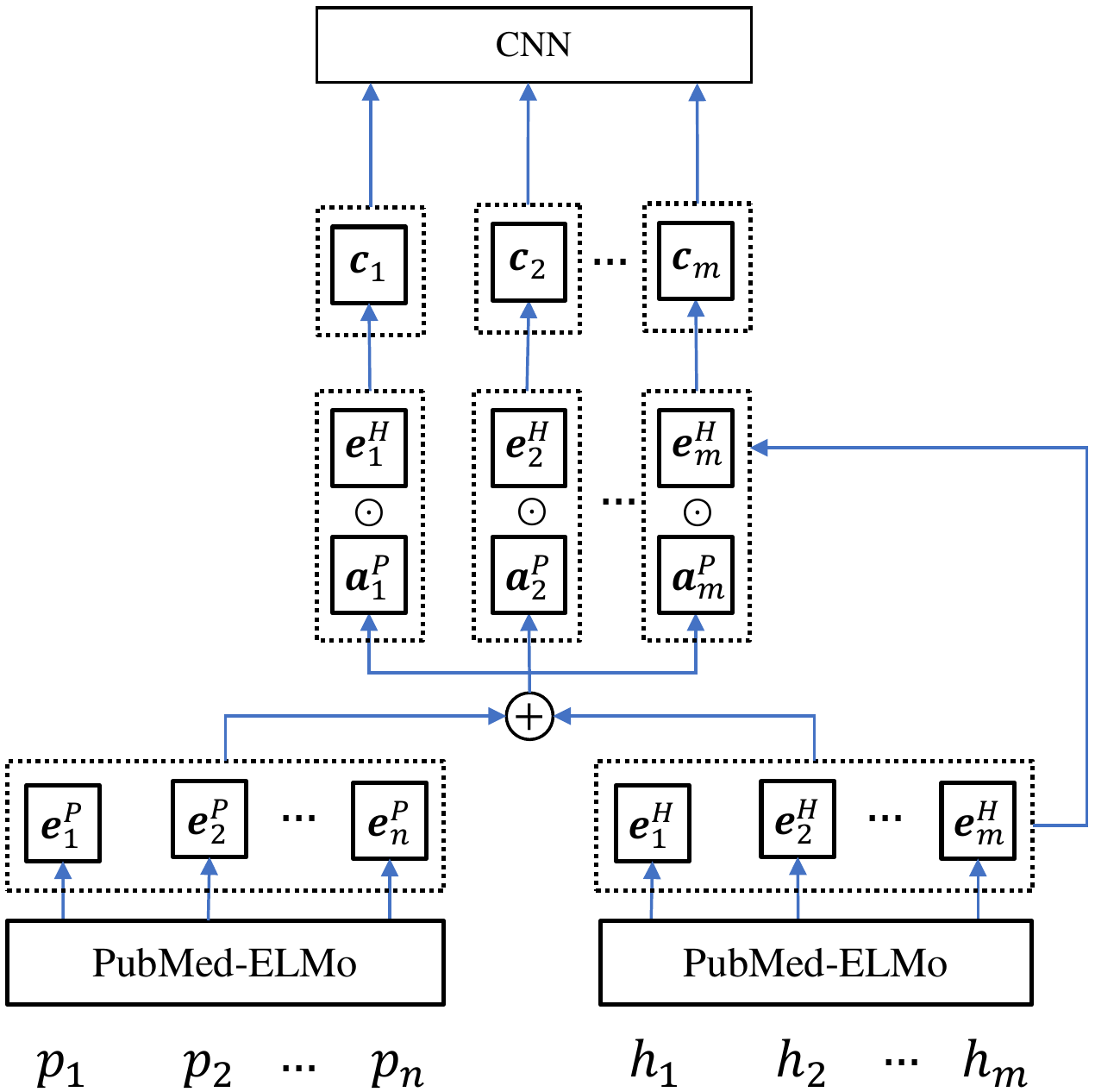}
\caption{
Overview of the CompAggr model.
}
\label{fig:comp_aggr}
\end{figure}

\subsection{Compare Aggregate (CompAggr)}
As we focus on the task that classifies the relationship between two sentences \textbf{P} and \textbf{H} (premise and hypothesis) into one of three classes (entailment, contradiction, or neutral), we adopt the compare aggregate (CompAggr) model that is widely used for a text sequence matching task~\cite{wang2016compare}. In addition to the CompAggr model, we adopt PubMed-ELMo, that is trained with medical domain corpus and released as one of contributed ELMo models~\cite{peters2018deep,pubmed_emlo2018}, to alleviate the lack of training corpus for the shared task. The final model consists of four parts which are shown in figure~\ref{fig:comp_aggr}.

\vspace*{1mm}
\noindent\textbf{Word representation: }
Premise $\textbf{P}\,{\in}\,\mathbb{R}^{d\times n}$ and hypothesis $\textbf{H}\,{\in}\,\mathbb{R}^{d\times m}$, (where \textit{d} is a dimensionality of word embedding and \textit{n}, \textit{m} are length of the sequences in \textbf{P} and \textbf{H}, receptively), are processed to capture contextual information within the sentence by using pretrained PubMed ELMo~\cite{peters2018deep} as follows:
\begin{equation}
\begin{aligned}
& \textbf{E}^P=\text{PubMed-ELMo}(\textbf{P}), \\
& \textbf{E}^H=\text{PubMed-ELMo}(\textbf{H}).
\end{aligned}
\label{eq:word_representation}
\end{equation}

\vspace*{1mm}
\noindent\textbf{Attention: }
The soft aliment of the $\textbf{E}^P$ and $\textbf{E}^H$ are computed by applying an attention mechanism over the column vector in $\textbf{E}^P$ for each column vector in $\textbf{E}^H$. Using an attention weight $\alpha_{i}$ for each column vector in $\textbf{E}^P$, we obtain a corresponding vector $\textbf{A}^P\,{\in}\,\mathbb{R}^{d\times m}$ from weighted sum of the column vectors of $\textbf{E}^P$.
\begin{equation}
\begin{aligned}
& \textbf{A}^P=\textbf{E}^P\cdot~\text{softmax}({(\textbf{W}{\textbf{E}^P})}^{\intercal}\textbf{E}^H),\\
\end{aligned}
\label{eq:attention}
\end{equation}
where $\textbf{W}$ is a learned model parameter matrix.

\vspace*{1mm}
\noindent\textbf{Comparison: }
We use an element-wise multiplication as a comparison function to combine each pair of $\textbf{A}^P$ and $\textbf{E}^H$ into a vector $\textbf{C}\,{\in}\,\mathbb{R}^{d\times m}$.

\vspace*{1mm}
\noindent\textbf{Aggregation: }
Finally \citet{kim2014convolutional}'s CNN with \textit{n}-types of filters is applied to aggregate all the information followed by another fully connected layer to classify the \textbf{P} and \textbf{H} pair as follow:
\begin{equation}
\begin{aligned}
& \textbf{R}=\text{CNN}(\textbf{C}),~~(R\,{\in}\,\mathbb{R}^{nd})\\
& \hat{y}_{c} = \text{softmax}((\textbf{R})^\intercal~\textbf{W}+\textbf{b}~), \\
\end{aligned}
\label{eq:aggregation}
\end{equation}
where $\hat{y}_{c}$ is the predicted probability distribution for the target classes and the $\textbf{W}\,{\in}\,\mathbb{R}^{nd\times 3}$ and bias \textbf{b} are learned model parameters.

Our loss function is cross-entropy between predicted labels and true-labels as follow:
\begin{equation}
\begin{aligned}
\mathcal{L} = -\log \prod_{i=1}^{N} \sum_{c=1}^{C} y_{i,c} \text{log} (\hat{y}_{i,c}),
\end{aligned}
\label{eq_loss}
\end{equation}
where $y_{i,c}$ is the true label vector, and $\hat{y}_{i,c}$ is the predicted probability from the softmax layer.
$C$ is the total number of classes (entailment, contradiction, and neutral for this task), and $N$ is the total number of samples used in training.

\subsection{Transfer learning}
\citet{pan2010survey} provide definitions of transfer learning as follows:

\vspace*{1mm}
\noindent\textbf{Definition 1} (\textit{Transfer Learning}) Given a source domain ${\cal D}_S$ and learning task ${\cal T}_S$, a target domain ${\cal D}_T$ and learning task ${\cal T}_T$, \textit{transfer learning} aims to help improve the learning of the target predictive function $f_T(\cdot)$ in ${\cal D}_T$ using the knowledge in ${\cal D}_S$ and ${\cal T}_S$, where ${\cal D}_s \not= {\cal D}_T$, or ${\cal T}_S \not= {\cal T}_T$.

While MedNLI has a relatively large amount of training data in the clinical domain, NLI tasks in general domain such as SNLI~\cite{bowman2015large} and MNLI~\cite{williams2018broad} have way larger training data than MedNLI has.
Since a source and a target task in different domains can improve a model performance if they are related to each other we decide to use the two general domain NLI tasks to train BERT and BioBERT to transfer their knowledge for MedNLI.
Our case is ${\cal D}_S \not= {\cal D}_T$ where the feature spaces between the domains are different or the marginal probability distributions between domain datasets are different ($P(X_S) \not= P(X_T)$).


\begin{table}[t]
\centering
\begin{tabular}{L{0.47\columnwidth}C{0.17\columnwidth}C{0.17\columnwidth}}
\toprule
\multirow{2}{*}{\textbf{Dataset}} & \multicolumn{2}{c}{\textbf{Accuracy}} \\
\cmidrule{2-3}
                         & dev        & test       \\
\midrule
+PMC & 80.50  & 78.97 \\
+PubMedd    & 81.14 & 78.83 \\
+PubMed+PMC & \textbf{82.15} & \textbf{79.04} \\

\bottomrule
\end{tabular}
\caption{
The BioBERT performance on the MedNLI task. Each model is trained on three different combinations of PMC and PubMed datasets (top score marked as bold).
}
\label{table:biobert_performance}
\end{table}







\subsection{Abbreviation expansion}
Not unlike other medical text, abbreviations and acronyms are easily found throughout the text in MedNLI as table~\ref{table:mednli_examples} shows from \# 4 to 6. In order to understand the effect of expanded forms for clinical abbreviations, we replace the abbreviations with corresponding expanded forms. As \citet{liu2015exploiting} mentions that no universal rules or dictionary for clinical abbreviations is available we gather and exploit the public medical abbreviations from Taber's Online\footnote{https://www.tabers.com/tabersonline/view/Tabers-Dictionary/767492/all/Medical_Abbreviations}.


\section{Experiments}
\label{sec:experiments}
We explore three kinds of BioBERT that are fine-tuned from the original BERT with PMC, PubMed, and PMC+PubMed datasets.
As shown in table~\ref{table:biobert_performance}, BioBERT trained on PubMed+PMC performs the best. Thus we select it as a base BioBERT model for the rest of the experiments.
Depends on a need for comparison or better understanding, we also include original BERT in the experiments and report the results.
The overall results of MedNLI are shown in table~\ref{table:overall_performance}.

\subsection{Experimental Setup}
All experiments based on BioBERT and BERT have a fixed learning rate 2e-5. We add early stopping to stop the models from learning if evaluation loss has not decreased for 4 steps where 1 step is defined 20\% of the whole training data. Other than the learning rate and early stopping, all settings are the same as they are in BioBERT and BERT.

For the CompAggr model, we use a context projection weight matrix with 100 dimensions. In the aggregation part, we use 1-D CNN with a total of 500 filters, which involved five types of filters $K\,{\in}\,\mathbb{R}^{\{1,2,3,4,5\}\times100}$, 100 per type. The weight matrices for the filters were initialized using the Xavier method \cite{glorot2010understanding}.
We use the Adam optimizer~\cite{kingma2014adam} including gradient clipping by norm at a threshold of 5.
For the purpose of regularization, we applied dropout~\cite{srivastava2014dropout} with a ratio of 0.7.

\begin{table}[t]
\centering
\begin{tabular}{L{0.47\columnwidth}C{0.17\columnwidth}C{0.17\columnwidth}}
\toprule
\multirow{2}{*}{\textbf{Model}} & \multicolumn{2}{c}{\textbf{Accuracy}} \\
\cmidrule{2-3}
                         & dev        & test       \\
\midrule
BioBERT    &  82.15  & 79.04 \\
CompAggr          &  80.40         & 75.80 \\
BioBERT (transferred)  & 83.51      & \textbf{82.63} \\
BioBERT (expanded)     & \textbf{83.87}      & 79.95  \\

\bottomrule
\end{tabular}
\caption{The model performance of four different methods (top score marked as bold). BioBERT (transferred) and BioBERT (expanded) refer to the best results of transfer learning experiments and the result of MedNLI with abbreviation expansion on BioBERT respectively. 
}
\label{table:overall_performance}
\end{table}


\begin{table*}[t]
\centering
\begin{tabular}{L{0.70\columnwidth}C{0.25\columnwidth}C{0.25\columnwidth}C{0.25\columnwidth}C{0.25\columnwidth}}

\toprule
\multirow{2}{*}{\textbf{Dataset}} & \multicolumn{2}{c}{\textbf{BERT}} & \multicolumn{2}{c}{\textbf{BioBERT}} \\
\cmidrule{2-5}
 & dev & test &  dev &   test\\
\midrule
MedNLI & 79.56 & 77.49 & 82.15 & 79.04 \\
MNLI (M) & 83.52 & - & 81.23 & -  \\
SNLI (S) & 90.39 & - & 89.10 & - \\

\midrule
M $\to$ MedNLI & 80.14 & \textbf{78.62} & 82.72 & 80.80 \\
S $\;\to$ MedNLI & 80.28 & 78.19 & 83.29 & 81.29   \\

\midrule
M $\to$ S $\to$ MedNLI &  80.43 & 78.12 & 83.29 & 80.30  \\
S $\to$ M $\to$ MedNLI &  \textbf{81.72} & 77.98 &  \textbf{83.51} & \textbf{82.63}  \\

\midrule
MedNLI (expanded) & 79.13 &  77.07 & \textbf{83.87} &  79.95  \\
S $\to$ M $\to$ MedNLI (expanded) & \textbf{82.15} & \textbf{79.95} & 83.08 & \textbf{81.85}  \\

\bottomrule
\end{tabular}
\caption{All experiment results of transfer learning and abbreviation expansion (top-2 scores marked as bold). MedNLI (expanded) denotes MedNLI with abbreviation expansion.}
\label{table:all}
\end{table*}

\subsection{Performance evaluation}

\vspace*{1mm}
\noindent\textbf{Transfer learning: }
We conduct transfer learning on four different combinations of MedNLI, SNLI, and MNLI as it shown in the table~\ref{table:all} (line 4 to 7) and also add the results of general domain tasks (MNLI, SNLI) for comparison. As expected, BERT performs better on tasks in the general domain while BioBERT performs better on MedNLI which is in the clinical domain.  

In overall, positive transfer occurs on MedNLI. There are three things we can observe from the results. First of all, even though BioBERT is fine-tuned on general domain tasks before MedNLI, transfer learning shows better results than that fine-tuned on MedNLI directly. It implies that the same tasks in different domains have overlapping knowledge and transfer learning between the tasks effects positively on each other as the definition of transfer learning mentions in section~\ref{sec:methods}.
Second, the domain specific language representations from BioBERT are maintained while fine-tuning on general domain tasks by showing that the transfer learning results of MedNLI on BioBERT have better performance than the results on BERT (line 4 to 7).
Lastly, the accuracy of MNLI and SNLI on BioBERT is lower than the accuracy on BERT. The lower accuracy indicates that BioBERT captures different features such as medical terms and generate different representations than what BERT does which are helpful for the clinical domain task, MedNLI, but not for the other two tasks.

The best combination is SNLI $\to$ MNLI $\to$ MedNLI on BioBERT. We refer to the best result of transfer learning as BioBERT (transferred).


\begin{figure}[t]
\small
\centering
\includegraphics[width=0.95\columnwidth]{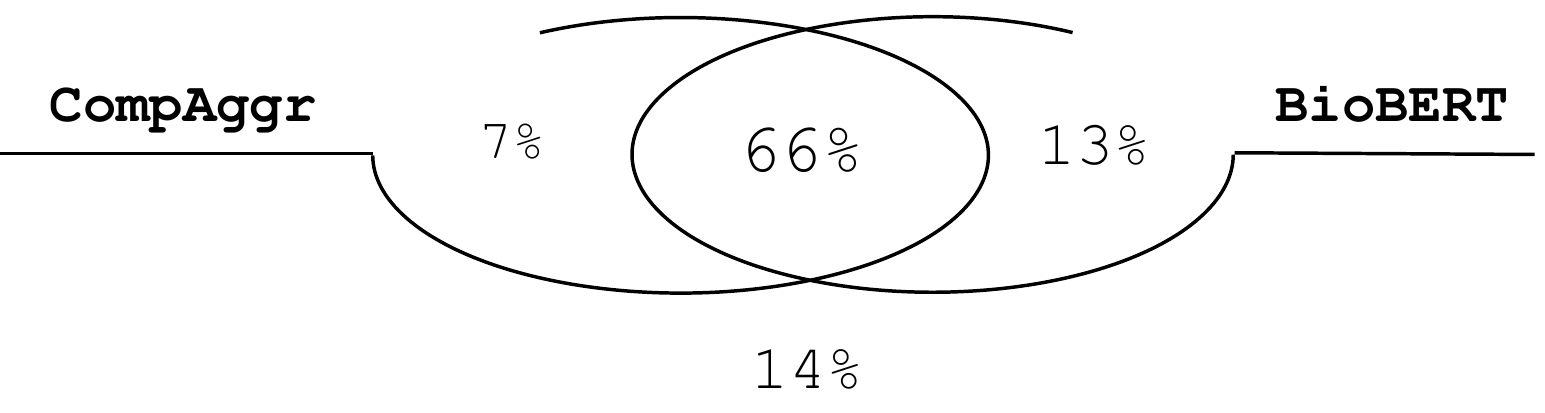}
\caption{
Venn diagram for the test results of Compare Aggregate model and BioBERT.
}
\label{fig:venn_compaggr_biobert}
\end{figure}

\vspace*{1mm}
\noindent\textbf{Results analysis for different models: }
There are fundamental differences between the two models we apply. BioBERT tokenizes an input sentence to sub-word level and uses the transformer model while CompAggr uses word-level embeddings and Compare\&Aggregate model. In light of the dissimilar nature, we expect each model captures different features and generates different language representations.

Figure~\ref{fig:venn_compaggr_biobert} shows the percentage for each area takes of the test set. CompAggr correctly classifies 97 examples (7\% of the test set) which BioBERT classifies them incorrectly while BioBERT classifies 188 examples correctly (13\% of the test set) which CompAggr does not. It demonstrates that both models have different strength on the MedNLI task.

\begin{table*}[t]
\centering
\small
\begin{tabular}{L{0.77\columnwidth}L{0.53\columnwidth}C{0.24\columnwidth}C{0.24\columnwidth}}
\toprule
 \textbf{Premise} & \textbf{Hypothesis} & \textbf{CompAggr} & \textbf{BioBERT}\\

\midrule
He denies any fever, diarrhea, chest pain, cough, URI symptoms, or dysuria.& He denies any fever, diarrhea, chest pain, cough, URI symptoms, or dysuria. & entailment & neutral \\

\midrule
This quickly became ventricular fibrillation and he was successfully shocked X 1 360J with return of rhythm and circulation.&  Patient has NSR post-cardioversion & entailment & contradiction\\

\midrule
PAST MEDICAL HISTORY: Coronary artery disease status post MI [**09**] years ago, status post angioplasty.& History of heart attack &entailment &neutral\\

\midrule
A MRA prior to discharge showed increased ... of single and rector spinal muscles at T3-4 adjacent to facets and anterior within the right psoas. & the patient has degenerative changes of the spine&entailment &neutral\\

\midrule
The patient now presents with metastatic recurrence of squamous cell carcinoma of the right mandible with extensive lymph node involvement. & The patient has oropharyngeal carcinoma.  &entailment &neutral\\

\midrule
The transbronchial biopsy was nondiagnostic. & Patient has a mediastinal mass &entailment &neutral\\

\bottomrule
\end{tabular}
\caption{Examples with the highest probabilities showing the strength of CompAggr. 
}
\label{table:compaggr_rlts}
\end{table*}
\begin{table}[t]
\small
\centering
\begin{tabular}{C{0.15\columnwidth}C{0.4\columnwidth}C{0.25\columnwidth}}
\toprule
\textbf{Rank} & \textbf{Team}  & \textbf{Accuracy} \\
\midrule
1 & WTMED & 98.0 \\
2 & PANLP & 96.6 \\
3 & Double Transfer & 93.8 \\
4 & Sieg & 91.1 \\
\textbf{5} & \textbf{Surf (ours)} & \textbf{90.6} \\
6 & ARS\_NITK & 87.7 \\
7 & Pentagon & 85.7 \\
8 & Dr.Quad & 85.5 \\
9 & UU\_TAILS & 85.2 \\
10 & KU\_ai & 84.7 \\
\bottomrule
\end{tabular}
\caption{
Performance comparison among the top-10 participants (official) of the NLI shared task. Teams [1-4, 6-10] are from~\cite{wtmed,panlp,doubletransfer,sieg,arsnitk,pentagon,drquad,uutails,kuai}, respectively.
}
\label{table:leaderboard}
\end{table}

We manually examine all promise and hypothesis pairs of each portion of 7\% and 13\% of the test set with high confidence and ``element" label. For CompAggr, we pick pairs with the probability higher than 0.80 which are 6 pairs. For BioBERT, we select pairs with top 10 probabilities. Interestingly, each pair from CompAggr does not have overlapping words between premise and hypothesis. It appears that CompAggr's strength is in it's ability to capture the relationship between two sentences even though there is no word overlap
while BioBERT labels them ``\textit{neutral}" except one pair as you can see in table~\ref{table:compaggr_rlts}. In contrast, the majority of the pairs, 7 out of 10, from BioBERT have overlapping words between them. Biobert shows strong confidence when premise and hypothesis have overlapping words as below.
\begin{itemize}
\item (\textit{Premise}) En route to the Emergency Department, she developed worsening substernal chest pain without any radiation.
\item (\textit{Hypothesis}) patient has chest pain
\end{itemize}

Lastly, we compute the average conditional probability of the correct results to check the confidence of each model. The results are 0.87 and 0.82 for BioBERT and CompAggr showing that BioBERT predicts labels with higher confidence.



\vspace*{1mm}
\noindent\textbf{Abbreviation expansion: }
We refer to the dataset of MedNLI with abbreviation expansion as MedNLI (expanded). The inconsistency of the experiment results on MedNLI (expanded) makes it difficult to observe their effects. 
MedNLI (expanded) shows better performance than MedNLI on BioBERT while MedNLI works better on BERT  (see table~\ref{table:all}). Furthermore, the performance of MedNLI (expanded) with transfer learning is higher on BERT and lower on BioBERT than the performance of MedNLI with transfer learning.

We examine the test results to figure out the inconsistency and observe an interesting phenomenon that the abbreviation expansion changes the conditional probability distribution P(Y$|$X), where X and Y represent input texts and their expected labels, respectively. The same input texts with no expansion are classified into different classes. For instance, a pair of {\em Premise} and {\em Hypothesis} like below is not changed after abbreviation expansion since it does not contain any abbreviations or acronyms. 
\begin{itemize}
\item (\textit{Premise}) He denied headache or nausea or vomiting.
\item (\textit{Hypothesis}) He is afebrile.
\end{itemize}
However, the results are different. It is originally classified into ``\textit{neutral}" which is the right label for the pair but it is classified into ``\textit{entailment}" when we use MedNLI (expanded).




\subsection{MEDIQA-NLI shared task}
\label{ssec:mediqa}
We are participating in a shared task MEDIQA-NLI of the bioNLP workshop at ACL 2019. In order to solve the task, we try four different \textbf{point-wise} approaches, CompAggr, BioBERT, transfer learning, and abbreviation expansion. We run each model several times to obtain the best result out of each. Our best result, which is ranked 5th on the leaderboard of the task, is obtained by applying \textbf{list-wise} approach (in section~\ref{sec:dataset}) with the best result (BioBERT (transferred)). Table~\ref{table:leaderboard} shows the model performance of each participant in the leaderboard. 




\section{Conclusion}
\label{sec:conclusion}
In this paper, we study natural language inference in the clinical domain where training corpora is insufficient due to its domain nature. 
To tackle the problem, we propose approaches that adopts pre-trained language models, transfer learning method and data-augmentation to boost the train instances.
To this end, we observe that the BioBERT pre-trained on bio-medical corpus shows better performance than that of the BERT on the general domain corpus.
The CompAggr with bio-ELMO and the BioBERT behave differently in classifying the MedNLI dataset due to the difference in their own architecture. 
Transfer learning with NLI tasks in general domain, (MNLI, SNLI), does not hurt the ability of the BioBERT capturing language representations of the clinical domain. 
In addition, we observe that it transfers positive knowledge from general NLI tasks to the MedNLI task. 
In contrast, a abbreviation expansion method needs particular care when adopting since it may hurt the model to predict the conditional probability distribution of the task.


\section*{Acknowledgments}
We sincerely thank the reviewers for their in depth feedback that helped improve the paper. K. Jung is with Automation and Systems Research Institute (ASRI), Seoul National University, Seoul, Korea, and was supported by the Ministry of Trade, Industry \& Energy (MOTIE, Korea) under Industrial Technology Innovation Program (No.10073144).

\bibliography{acl2019}
\bibliographystyle{acl_natbib}

\end{document}